\documentclass[letterpaper, 10 pt, conference]{ieeeconf}  

\IEEEoverridecommandlockouts                              

\usepackage{graphicx, color, comment, amsmath, mathtools}

\usepackage{hyperref}
\hypersetup{
    colorlinks=true,
    linkcolor=black,
    filecolor=black,      
    urlcolor=black,
    pdftitle={Overleaf Example},
    pdfpagemode=FullScreen,
    }
    
\usepackage{microtype}
\renewcommand{\baselinestretch}{0.97}

\graphicspath{{images/}}

\title{\LARGE \bf
HAT: Head-Worn Assistive Teleoperation of Mobile Manipulators  
}

\author{Akhil Padmanabha$^{1*}$, Qin Wang$^{1*}$, Daphne Han$^{1}$, Jashkumar Diyora$^{1}$, Kriti Kacker$^{1}$, Hamza Khalid$^{1}$,\\ Liang-Jung Chen$^{1}$, Carmel Majidi$^{1,2}$, Zackory Erickson$^{1}$
\thanks{$^{*}$These authors contributed equally and are ordered alphabetically.}%
\thanks{This work was supported by the National Science Foundation Graduate Research Fellowship Program under Grant No. DGE1745016 and DGE2140739.}
\thanks{$^{1}$ Robotics Institute, Carnegie Mellon University, Pittsburgh, PA, USA}
\thanks{$^{2}$ Department of Mechanical Engineering, Carnegie Mellon University, Pittsburgh, PA, USA.}%
\thanks{Akhil Padmanabha is the corresponding author. \newline{\tt\small akhilpad@andrew.cmu.edu}}%
}

\begin{document}

\maketitle
\thispagestyle{empty}
\pagestyle{empty}

\begin{abstract}
Mobile manipulators in the home can provide increased autonomy to individuals with severe motor impairments, who often cannot complete activities of daily living (ADLs) without the help of a caregiver. Teleoperation of an assistive mobile manipulator could enable an individual with motor impairments to independently perform self-care and household tasks, yet limited motor function can impede one's ability to interface with a robot. In this work, we present a unique inertial-based wearable assistive interface, embedded in a familiar head-worn garment, for individuals with severe motor impairments to teleoperate and perform physical tasks with a mobile manipulator. We evaluate this wearable interface with both able-bodied (N = 16) and individuals with motor impairments (N = 2) for performing ADLs and everyday household tasks. Our results show that the wearable interface enabled participants to complete physical tasks with low error rates, high perceived ease of use, and low workload measures. Overall, this inertial-based wearable serves as a new assistive interface option for control of mobile manipulators in the home.%
\end{abstract}

\section{INTRODUCTION}
Motor impairments and loss of hand function can restrict an individual from performing activities of daily living (ADLs), such as eating, self hygiene, and dressing. Many individuals with lost motor function rely on assistance from caregivers, limiting their self-sufficiency, independence, and control in self-care and household tasks~\cite{qol1, qol2, qol3}. Motor impairments affect a significant subset of the United States with approximately 1.7\% of the population, over 5 million people, reportedly living with some form of paralysis~\cite{armour2016prevalence}. Loss of motor function can be due to a range of neurodegenerative diseases, stroke, muscle atrophy, and spinal cord injury (SCI). As an example, more than 100,000 individuals in the United States currently live without the ability to use their hands because of a cervical SCI~\cite{armour2016prevalence, scistat1, scistat2}. Many individuals with SCI lose some or all ability to use their upper and/or lower extremities, but often retain motion in their head and neck~\cite{world2013international}. For many with tetraplegia or limited hand function, technologies that restore the ability to perform everyday tasks can have an immediate and profound impact~\cite{impact1, impact2, impact3, impact4, impact5}. Teleoperation of mobile manipulator robots can enable individuals with motor impairments to once again perform physical tasks including self-care, fetching objects, etc.~\cite{tasks, tasksaroundhead, robotsforhumanity, king2012dusty}.  

While developing teleoperation interfaces for individuals with impairments, it is important to consider ease of use, efficiency (performing a task with a minimal number of steps in a short amount of time), and comfortability. Most interfaces have physical requirements, determining which subset of individuals with impairments will use them; for example, conventional hand-operated joysticks provide precise, continuous control but require users to have motor function in their hand. Meanwhile, web interfaces, the current convention for teleoperation, rely on the use of a device with a screen, requiring the ability of an individual to move a cursor and click on buttons through an assistive device, often requiring fine motor control. The development and evaluation of novel assistive interfaces for mobile manipulators could lead to alternatives for individuals with impairments who may have a difficult time accessing traditional systems. 

\begin{figure}[t!]
      \centering
      \includegraphics[width = \columnwidth]{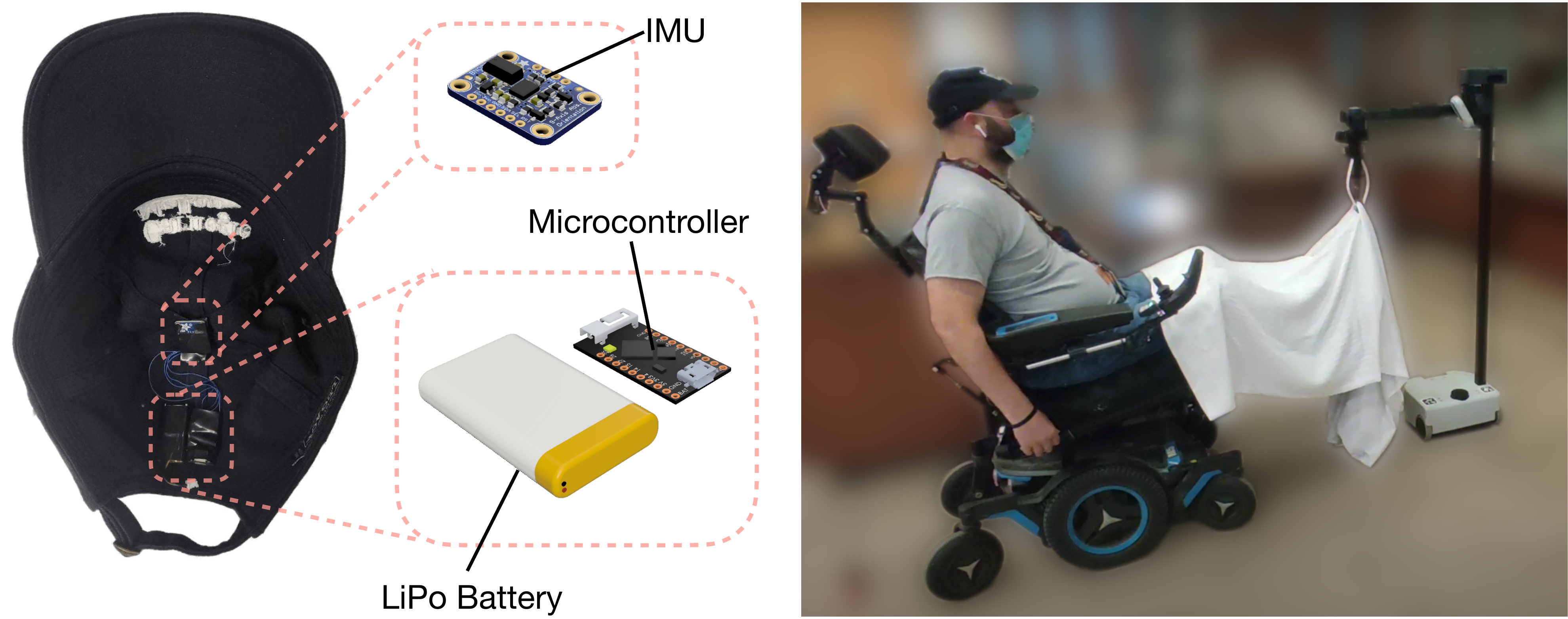}
      \vspace{-0.4cm}
      \caption{Left: The head-worn interface with integrated inertial measurement unit (IMU) sensing to control an assistive robotic manipulator. The hat consists of an absolute orientation IMU connected to a microcontroller with builtin Bluetooth capabilities and powered by a small lithium polymer (LiPo) battery. A thin layer of Neoprene foam is added over the electronics for comfort. Right: A participant (I1) with spinal cord injury completing a blanket removal task using the head-worn interface.}
      \vspace{-0.4cm}
      \label{hat}
   \end{figure}

In this work, we present a novel inertial-based assistive interface for individuals with motor impairments to directly control a lightweight mobile manipulator using residual head motion. The inertial-based interface, embedded in a baseball hat as shown in Fig \ref{hat}, sends absolute orientation angles of the user's head to a mobile manipulator. The angles are mapped to velocity commands for actuation of the robot's base, arm, and end effector. We evaluated the interface using both quantitative and qualitative metrics through a human study with 16 able-bodied participants and 2 participants with motor impairments. Participants use this interface to control a mobile manipulator and complete four self-care and household tasks. Our results show that the head-worn device is an efficient and intuitive control interface for a mobile manipulator, presenting individuals with motor impairments a strong alternative to conventional teleoperation platforms.
The contributions of this work is as follows: 
\begin{itemize}
    \item We introduce a novel assistive interface with inertial measurements integrated into an everyday clothing article.
    \item We present how this inertial head-worn interface can enable teleoperation of a high degree-of-freedom mobile manipulator using only residual head motion retained after cervical spinal cord injury (SCI) or tetraplegia. 
    \item We evaluated this wearable interface with 16 able-bodied participants and 2 participants with motor impairments and show that the head-worn device enabled participants to complete self-care and household tasks with low error rates, high perceived ease of use, and low workload measures.
\end{itemize}

Build instructions and code for this interface are open-sourced on the project website\footnote{ \url{https://sites.google.com/view/hat-teleop/home}}.
    
\section{Related Work}
Current assistive interfaces for wheelchairs and mobile robots include hand and mouth operated joysticks~\cite{fehr2000adequacy}, and web-based visual interfaces~\cite{robotsforhumanity, surrogates, TapoMaya} which can be used in conjunction with other assistive devices such as head tracking~\cite{headtracking, headtracking2, FaceMouse}, eye tracking~\cite{jacob1995eye, kaufman1993eye}, etc. These web interfaces enable control of robot motions and often incorporate sub-modules that allow users to select autonomous routines such as grasping of an object. Researchers have evaluated several iterations of these teleoperation platforms with both healthy participants and one participant with tetraplegia using a PR2 robot~\cite{tasks,tasksaroundhead,robotsforhumanity}. This interface enabled control of the robot for various tasks including grasping in cluttered environments, handing out candy, opening of a drawer and extracting an object, etc. Efforts have been made to make these interfaces more intuitive, by overlaying movement buttons directly over camera feeds and providing additional views of the environment, and to evaluate them with more individuals with motor impairments~\cite{surrogates,TapoMaya}. 

Researchers have previously explored teleoperation of robots using an IMU-based device that captures hand and arm movements~\cite{mobilerobotIMU, IMUglove}. While highly efficient, these interfaces cannot be used by individuals without intact limb movement. Prior work also includes head-worn IMU-based interfaces for control of electric wheelchairs~\cite{wheelchairIMU, wheelchair2, wheelchair3}, but has not shown generalization to mobile manipulators that have many additional degrees of freedom. 

Lastly, brain computer interfaces (BCI) show promise for efficient control of physical devices for individuals with motor impairments. Researchers have demonstrated control of a robotic arm for tetraplegics using embedded neural interfaces~\cite{BrainGate1}. Even though potentially applicable to a large population, these BCI techniques are often invasive and need to be extensively calibrated with each new user. As an alternative to BCI, EEG and EMG signals from skin-mounted electrodes are non-invasive and have been used to teleoperate simple robot motions based on residual myoelectric signals~\cite{emg1, emg2}. In contrast, a head-worn noninvasive assistive interface may be generally applicable to a wide population of individuals who have lost significant motor function below the neck, as is the case for many individuals with cervical SCI or tetraplegia due to neurodegenerative disease, stroke, or injury.

\section{Device Design}
\begin{figure}[t!]
      \centering
      \includegraphics[width = \columnwidth]{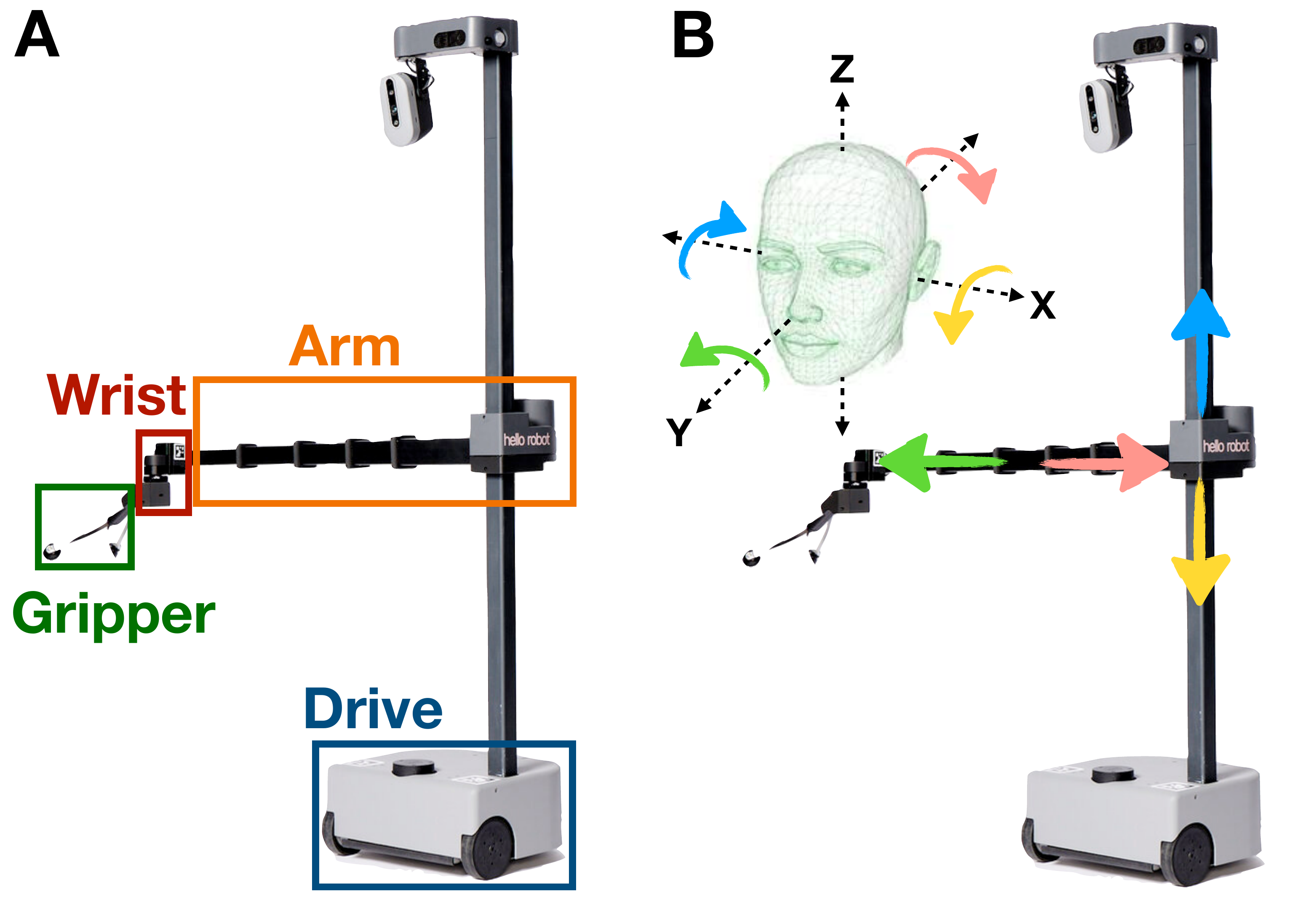}
      \vspace{-0.4cm}
      \caption{Mapping from the head-worn interface to the robot. (A) The 4 separate modes of the robot controlled by the interface. (B) Visual representation of the mapping from head movements to robot motion in the arm mode.}
      \vspace{-0.4cm}
      \label{robot}
   \end{figure}

In Fig.~\ref{robot}, signals from the head-worn interface are communicated to the mobile manipulator via Bluetooth and mapped to velocity commands for the robot's actuators. The interface is used for direct control of the robot's motion. Speech recognition, using audio captured by a wireless microphone worn by the participant, is used for selection of four robot modes: drive, arm, wrist, and gripper, as shown in Fig.~\ref{robot}A. The head movements are mapped to robot motions based on the mode the user is in; for example, the arm mode mapping is shown in Fig.~\ref{robot}B. 

The inertial head-worn interface, shown in Fig.~\ref{hat}, consists of an Adafruit BNO055 absolute orientation inertial measurement unit (IMU) which fuses accelerometer, gyroscope, and magnetometer data and outputs absolute roll, pitch, and yaw orientation angles. A Tiny PICO microcontroller is used to sample the IMU at 20 Hz and to send this data to the mobile manipulator over Bluetooth.  

\begin{figure*}[thpb]
      \centering
      \includegraphics[width = \textwidth]{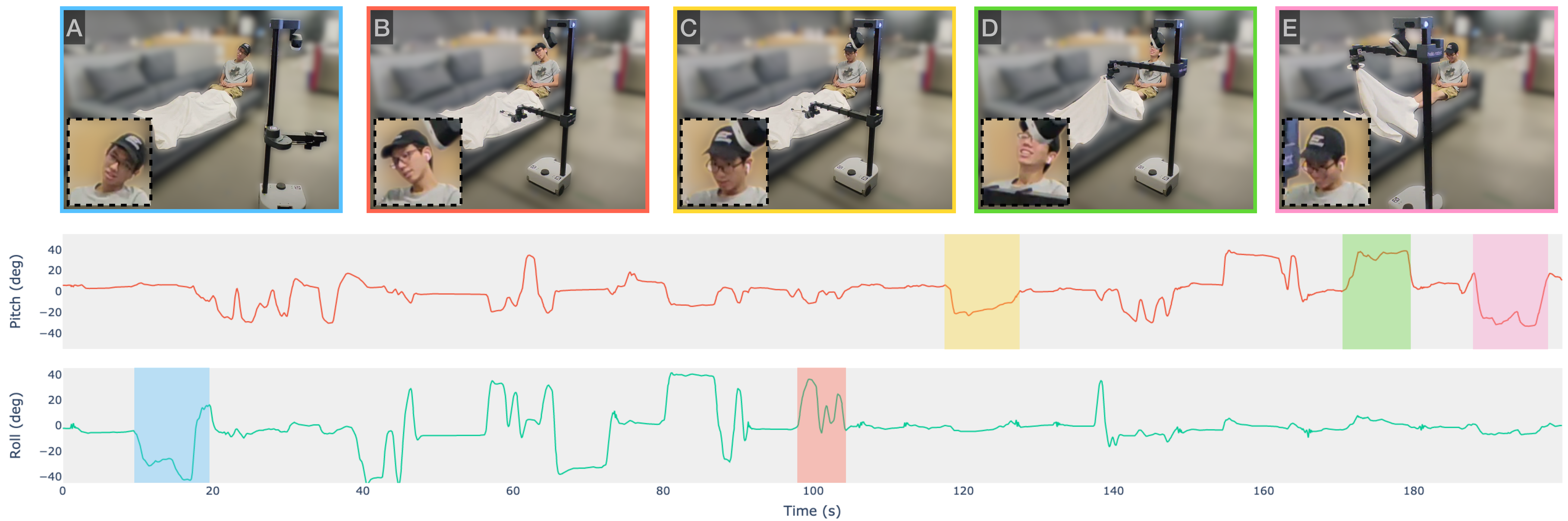}
      \vspace{-0.5cm}
      \caption{Time lapse of a participant dragging a blanket off his legs with corresponding signals from the IMU along the pitch and roll axes. The highlighted regions of the roll and pitch signals correspond to the matching colored video frames and correlate to the head motions that the participant performs to accomplish a specific robot motion. (A) Participant tilts their head to the left to make the robot turn towards their body. (B) After positioning the robot's base, the participant switches to arm mode and tilts their head to the right to extend the arm. (C) The participant switches to gripper mode and pitches their head down to open the gripper. (D) After grasping the blanket, the participant switches to arm mode and tilts their head up to raise the robot's arm up. (E) Lastly, the participant drives the robot forward by pitching their head forward.}
      \label{timelapse}
      \vspace{-0.5cm}
   \end{figure*}
   
\subsection{Mode Switching}
For the human study, participants use speech recognition to enable control of the robot by the head-worn interface and to switch between the four robot modes. While in each mode, users can command specific robot motions by tilting their head along the roll or pitch axis. The speech commands are ``start'', ``switch to drive'', ``switch to arm'', ``switch to wrist'', and ``switch to gripper''. To trigger the speech recognition, participants shake their head lightly to the right and left, along the Z-axis, shown in Fig~\ref{robot}B. The participants receive audible confirmation, either the recognized command or ``repeat'' if the phrase is unidentifiable. We foresee mode switching to be one of many plug-and-play options including a clicker, sip/puff device, or even head motions along the Z-axis based on a user's ability and preference. 

\subsection{Calibration and Robot Mapping}
The robot receives mode-switching commands and interface orientation angles over Bluetooth. The robot pushes commands to its actuators at 10 Hz. Before controlling the robot, users must calibrate the initial IMU orientation by triggering voice recognition and saying ``start''. During calibration, the orientation of the head-worn interface is saved and used for setting four thresholds. The minimum motion thresholds, $t_{l,a}$ and $t_{l,b}$, are set at $15^{\circ}$ and $-15^{\circ}$ respectively from the calibrated position, $\theta_c$. The maximum thresholds, $t_{h,a}$ and $t_{h,b}$, are set at $45^{\circ}$ and $-45^{\circ}$ respectively from the calibrated position, $\theta_c$. If the user's head is tilted less than $t_{l,a}$ and greater than $t_{l,b}$, in both the X and Y axes, the robot stops all motion. For a single axis (either X or Y), the angle measurement is proportionally scaled to velocity commands for the robot actuators according to the following equation:
\[ 
    V(\theta, \theta_c, a) = 
    \begin{cases} 
    -v_{a,max} & \text{if } t_{h,b} > \theta 
    \\
    k_a(\theta - t_{l,b}) &
    \text{if }  t_{l,b}  \geq  \theta \geq t_{h,b}
    \\
    0 & \text{if } t_{l,b}  < \theta < t_{l,a}
    \\
    k_a(\theta - t_{l,a}) &
    \text{if }  t_{l,a} \leq  \theta \leq  t_{h,a}
    \\
    v_{a,max} & \text{if } t_{h,a} < \theta 
   \end{cases}
\]
where $V(\theta, \theta_c, a)$ is the velocity command sent to actuator $a$, $\theta$ is the angle of the user's head, $\theta_c$ is the initial angle of the user's head during calibration, $k_a$ is the proportional constant for actuator $a$, $v_{a,max}$ is the maximum velocity limit for actuator $a$ , $t_{l,a} = 15^{\circ} + \theta_c$, $t_{h,a} = 45^{\circ} + \theta_c$, $t_{l,b} = -15^{\circ} + \theta_c$, and $t_{h,b} = -45^{\circ} + \theta_c$.

In drive mode, pitch of the head controls forward and backwards motion of the robot's base, whereas roll controls rotation clockwise and counterclockwise. In arm mode, shown in Fig.~\ref{robot}B, pitch controls the robot arm height and roll controls the extension. In wrist mode, pitch controls the wrist's pitch and roll controls the wrist's yaw. Lastly, in gripper mode, pitch commands the gripper open and close. The yaw (Z) axis of the IMU is only used for triggering speech recognition and not for any robot motion, allowing the user to rotate their head in this axis to look at objects in their environment. A time lapse of a participant completing a blanket task from the human study is shown in Fig.~\ref{timelapse} with corresponding pitch and roll angles from the head-worn interface. 

\section{Study Design}

\begin{figure*}[thpb]
      \centering
      \includegraphics[width = \textwidth]{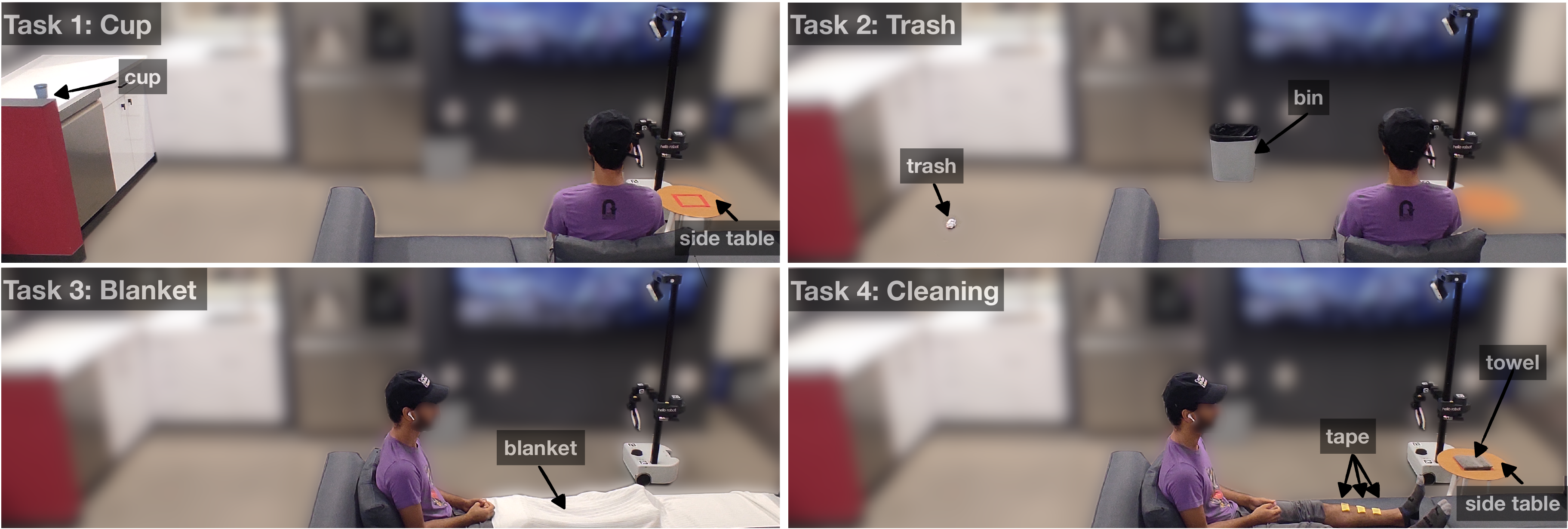}
      \vspace{-0.5cm}
      \caption{Setup of the 4 study tasks: Cup, Trash, Blanket, and Cleaning. The robot starts in the same position for each task, homed and with the wrist stowed. For Task 1 (Cup), participants must use the robot to grab a cup from a kitchen counter and place it within a square target on the side table. For Task 2 (Trash), participants are instructed to grab a piece of trash (crumpled A4 printer paper) on the floor, and place it in a trash bin. For Task 3 (Blanket), participants are told to remove the blanket from their legs using the robot. The task ends when the blanket is not touching any part of the participant. Finally, for Task 4 (Cleaning), participants must grab a towel from the side table and use it to clean 3 pieces of ``dirt'' (paper tape pieces) stuck to their left leg. The task ends when all pieces of tape are unstuck and moved from their original position.}
      \vspace{-0.3cm}
      \label{tasks}
   \end{figure*}

The human study was conducted with 16 healthy participants (11 M, 5 F) with ages ranging from 18-27 (Mean = 22.8, STD = 2.5) and two participants with motor impairments. We obtained informed consent from all participants and approval from the Carnegie Mellon University Institutional Review Board. Participants were asked how much experience they have with controlling a robot on a 5-point scale ranging from ``1 = No Experience'' to ``5 = Expert User''. Participants had a mean experience of 2.2 (STD = 1.2, Median = 2).   

The participants are instructed to do 4 tasks (cup retrieval, trash pickup, blanket removal, and leg cleaning / hygiene) by controlling the robot in a replicated home environment using the head-worn interface. The tasks are ordered in difficulty and shown visually in Fig~\ref{tasks}. Participants were allowed to ask researchers to reset the environment in cases when an object is dropped or they think it would be easier to restart; however, the task timer was not reset. Participants were given 14 minutes to complete each task and are instructed to keep their body still and only move their head and neck.

The head-worn interface serves as an alternative to a web interface for people who have a difficult time accessing traditional computing systems. 
Thus, for Task 1 (Cup), we additionally have participants use the modified Stretch web interface with modifiable speed control~\cite{TapoMaya} as a baseline. The order of the web and head-worn interface for Task 1 was alternated for every participant. For the web interface, participants used a laptop placed in front of them with a head tracking software\footnote{Enable Viacam (\url{https://eviacam.crea-si.com/index.php})} to move the cursor and a single button mouse to click. Head tracking software is a common interface used by many individuals with head control but limited hand function to interact with computational devices~\cite{headtracking, headtracking2, FaceMouse}. All healthy participants used head tracking software with the web interface to standardize control between the head-worn and web interfaces. 

Prior to the study, participants are led through practice sessions for both interfaces. Participants watched an instructional video for each interface\footnote{HAT Instructional Video (\url{https://youtu.be/v8wXM-cCss0})} and were led through a series of commands available. For the web interface, participants watched the instructional video from~\cite{TapoMaya}. Finally, participants completed a practice task, grabbing a wooden block on a kitchen counter, while the researchers provided tips. After the practice sessions, participants are not provided any help with using the interfaces. 

After each completed task, participants answer two 7-point Likert items provided in Fig.~\ref{lik12}. At the end of the study, participants answered the six question NASA Task Load Index (TLX), which measures perceived workload, and 7-point Likert items provided in Fig.~\ref{finlik}. A researcher also asks participants the following open-ended questions for each interface: (1) What did you like about the interface? (2) What were the cons of the interface? (3) What were the reasons for any errors you encountered? (4) What were the biggest challenges while using the interface? Lastly, participants are asked how the two interfaces compare to each other.

\subsection{Participants with Motor Impairments}
The procedure described above was additionally followed for two participants with impairments, who were given the choice to conduct the study either in the same location as healthy participants or in an alternate location of their preference. Participants were allowed to use a device of their choosing for controlling the web interface and had no time limits to complete each task. 

Participant I1 (M, 36) is paralyzed below the waist and additionally has limited movement in his hands and neck due to spinal cord injury. He is an expert robot user (marked 5 on survey) and has participated in robotics studies before. He chose to use a computer mouse for the web interface as he has enough hand control to do so, but uses his hand to operate the mouse in a non-standard manner. He conducted the study in an alternate location.

Participant I2 (M, 63) has limited hand function due to an essential tremor, a neurological disorder that causes involuntary shaking primarily while writing and using a keyboard. He has no prior experience with controlling a robot (marked 1 on survey). He chose to use a computer mouse for the web interface, but mentioned occasionally having hand tremors while doing so. He conducted the study from the same location as healthy participants.

\section{Results and Discussion}

\subsection{Task Completion and Efficiency}

\begin{figure}[t!]
  \centering
  \includegraphics[width = \columnwidth]{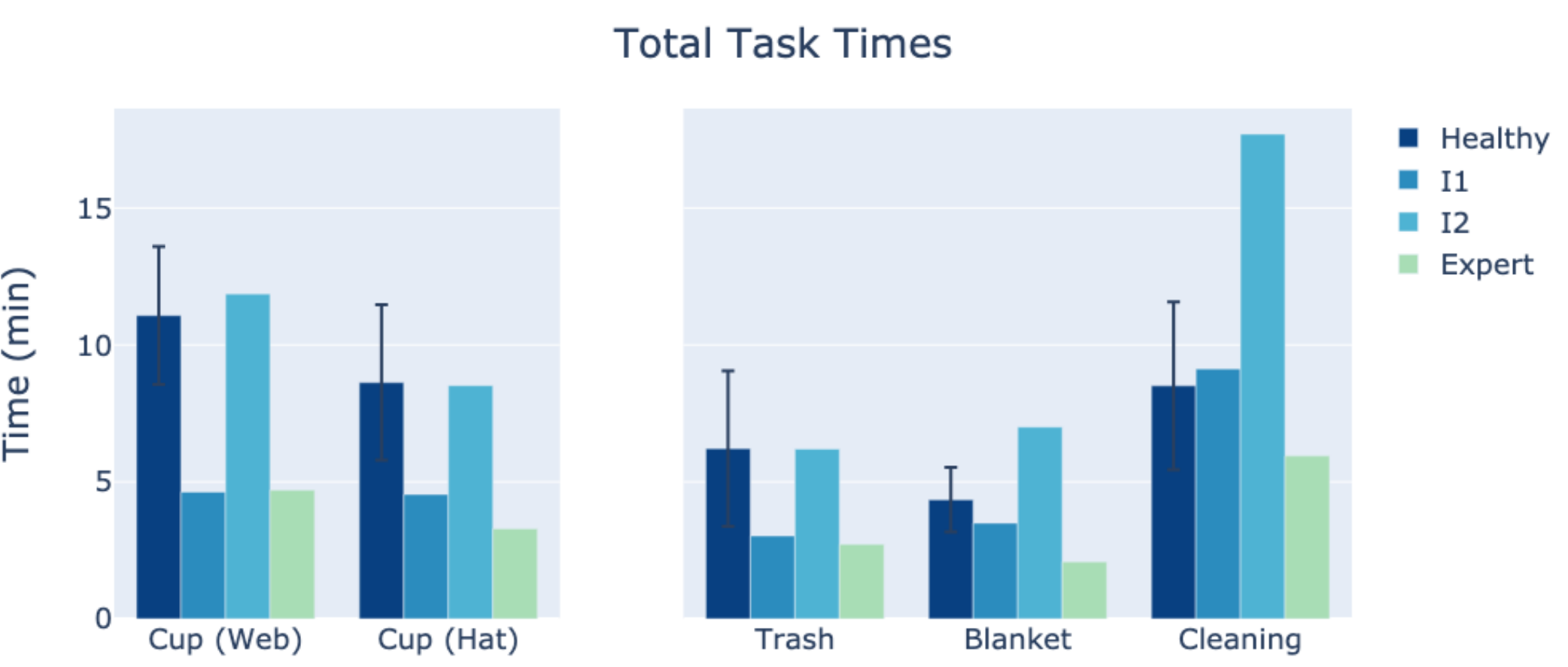}
  \vspace{-0.5cm}
  \caption{Average task times from 16 healthy participants, 2 participants with impairments, and 1 expert user. 4 healthy participants were unable to finish Task 1 (Cup) with the web interface due to reaching the time limit and 1 healthy participant additionally failed to complete Task 2 (Trash) and Task 4 (Cleaning) in the allotted time. If a task is not completed, the task time is recorded as 14 minutes.}
  \vspace{-0.5cm}
  \label{times}
\end{figure}

\begin{figure}[t!]
  \centering
  \includegraphics[width = \columnwidth]{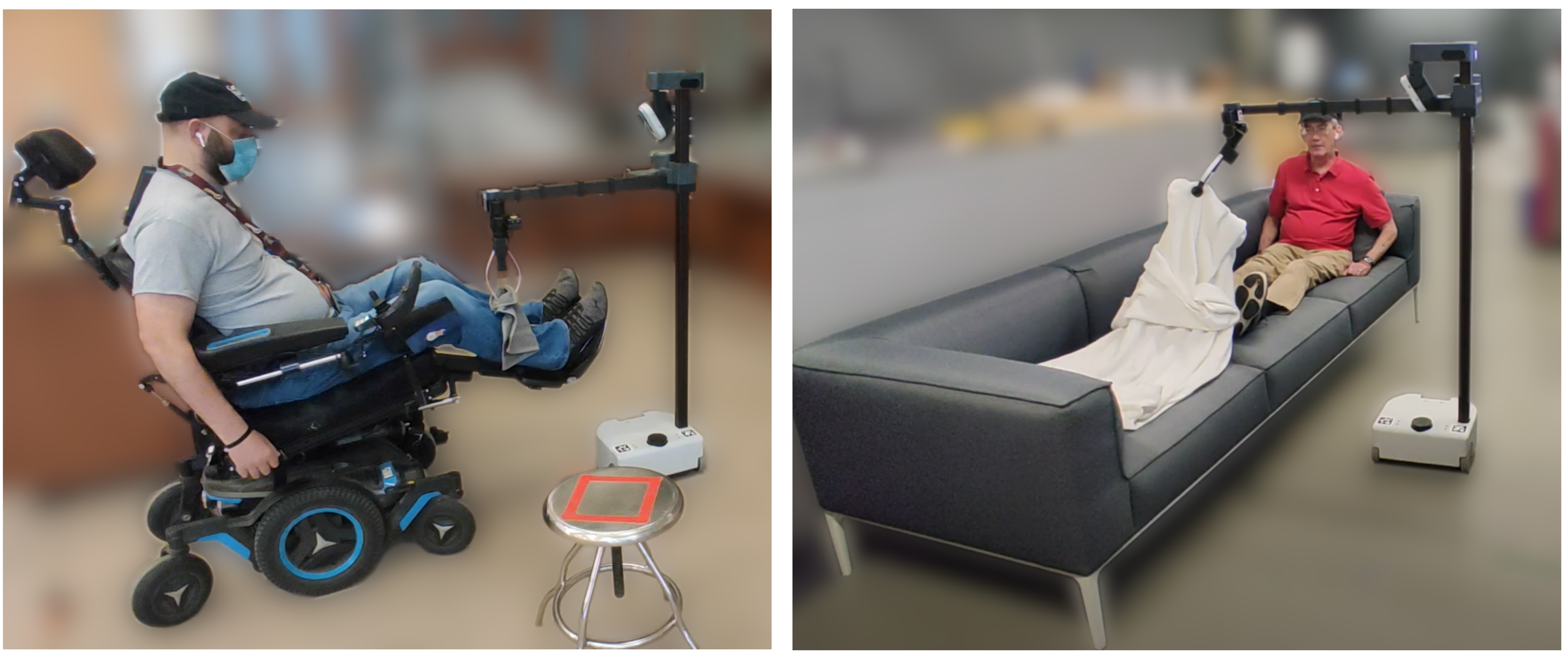}
  \vspace{-0.5cm}
  \caption{Left: Participant I1 completing Task 4 (Cleaning) using the head-worn interface. Right: Participant I2 completing Task 3 (Blanket) using the head-worn interface.}
  \vspace{-0.4cm}
  \label{impaired}
\end{figure}

The vast majority of participants in both groups successfully completed all head-worn interface tasks within the allotted time. Participant I1 and I2 are shown doing tasks from the study in Fig.~\ref{impaired}. Task completion times are shown in Fig.~\ref{times}. For Task 1 (Cup), the proposed head-worn interface allowed for faster task completion in comparison to the web interface with 12 out of 16 healthy participants and both participants with impairments finishing Task 1 faster. On average, Task 1 with the head-worn interface took 147 seconds less than with the web interface for healthy participants as shown in Fig.~\ref{times}. When asked the Likert item ``I was able to complete this task efficiently using the control interface'', shown in Fig.~\ref{lik12}, 15 out of 16 healthy participants and both participants with impairments ranked the head-worn interface the same as the web interface or higher, with a median ranking of 6 (Agree) and 4 (Neutral) by each group respectively. By applying a Wilcoxon signed-rank test, we observed a statistically significant difference between task times (p = 0.04) and between Likert item responses on task efficiency (p = 0.0016) for the head-worn and web interfaces on Task 1 with healthy participants. The head-worn interface was more efficient as it allowed participants to have continuous control of robot motion and speed (i.e. degree of head tilt resulting in more/less speed), thus requiring less commands than the web interface, which only allowed clicks of constant distances and selection of speed through buttons. 
 
As displayed in Fig.~\ref{lik12}, participants agreed with the Likert item on task efficiency for the three remaining tasks with the head-worn interface. Note that participants with motor impairments were able to complete most of the tasks in similar times to healthy participants as shown in Fig.~\ref{times}. Task 4 (cleaning) was the most challenging for participants (both healthy and with impairments) due to the added difficulty of wiping off tape that was firmly attached to their leg. Despite the increased task time, all participants were successful in grabbing the towel from the side table and wiping their leg. Additionally seen in Fig.~\ref{times} are the times achieved by an expert user, a member of the research team, for performing the same tasks; these results serve as a baseline for what task times are achievable with sufficient experience with the interface. At the end of the study, all healthy participants agreed that ``The [head-worn] interface enabled control of the robot in a reasonable amount of time'' with a median score of 6 (Agree) and both participants with impairments reported 7 (Strongly Agree), as detailed in Fig.~\ref{finlik}. 
 
\subsection{Errors and Recovery}

\begin{figure}[t!]
  \centering
  \includegraphics[width = \columnwidth]{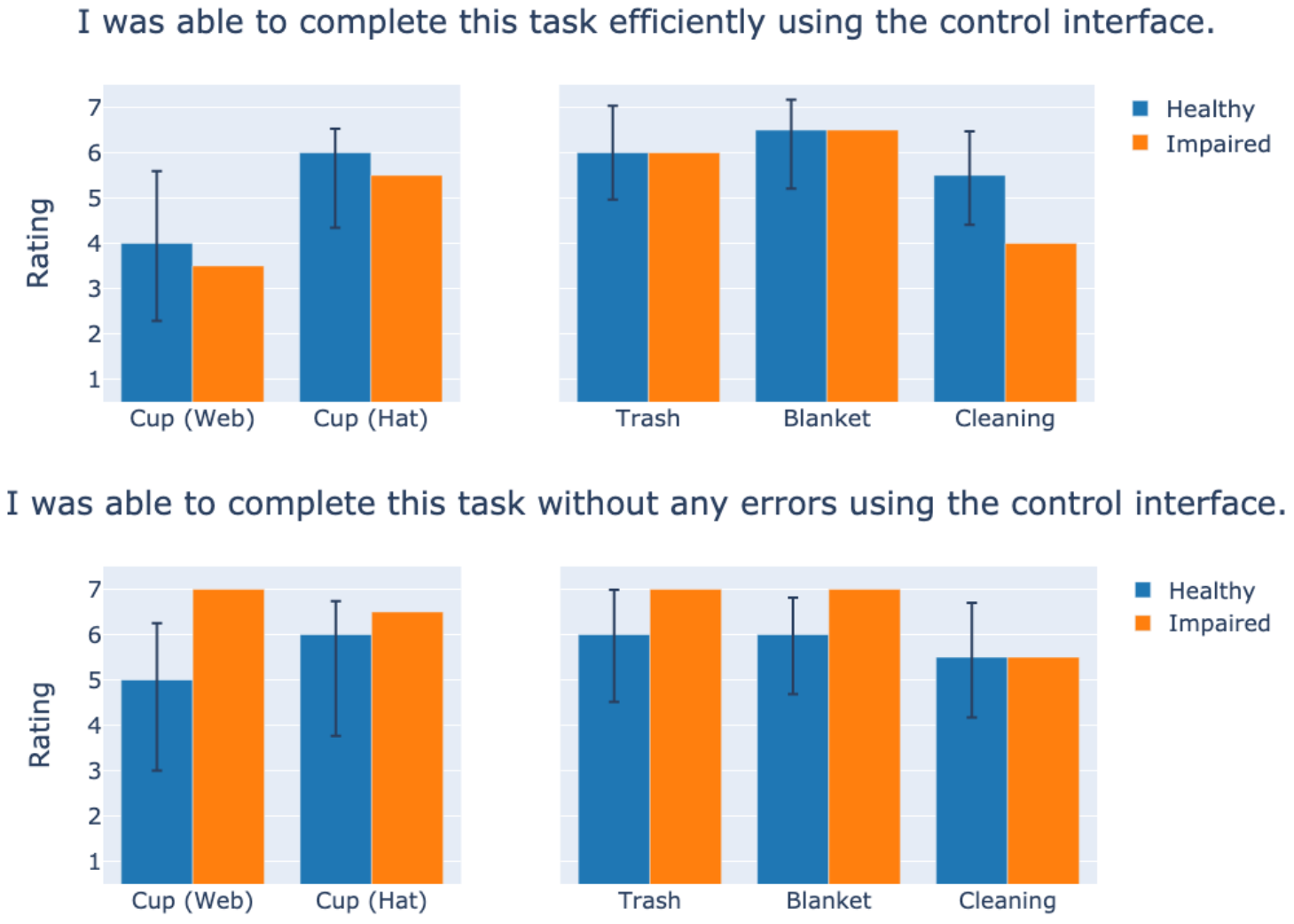}
  \vspace{-0.5cm}
  \caption{Median results from the two 7-point Likert items that participants are asked to evaluate after completion of each task.}
  \vspace{-0.5cm}
  \label{lik12}
\end{figure}

After each task, when asked the Likert item ``I was able to complete this task without any errors using the control interface'', median results show that both healthy participants and participants with motor impairments agreed for all  head-worn interface tasks, as shown in Fig.~\ref{lik12}. Nonetheless, some errors were observed with participants operating the head-worn interface. Common errors included participants tilting their head too far resulting in the robot moving too fast and overshooting the intended position. Occasionally, participants tilted their head the opposite direction of their intended action; I2 summarized this well saying, ``Biggest thing and it's not unique to this interface is keeping straight which direction is which when you're not oriented the same way the vehicle is.''. In contrast, 3 healthy participants and 1 participant with impairments mentioned liking the buttons of the web interface which would need to be clicked to cause robot motion, thus resulting in less inadvertent movements of the robot. For example, P11 mentioned ``I had to click the button to do something. [There was a] low chance of any accident.''. In most cases, these mistakes with the head-interface were quickly recognized by the participants and were easy to recover from. With regards to overshooting, 2 healthy and both participants with impairments mentioned similar downsides with the web interface as they were unable to predict how far the robot would move with each click of the web interface, especially after changing control speeds. For the head-worn interface, a calibration procedure, visual cues, and additional practice could help address these common sources of error identified through the human study. At the end of the study, we presented the following 7-point Likert item to participants regarding the head-worn interface: ``The control interface allowed easy recovery from errors''. As shown in Fig.~\ref{finlik}, healthy participants agreed with a median ranking of 5.5 while participants with impairments agreed to the statement with a median response of 6.5 (out of 7).

\subsection{Ease of Use}
Another important factor to consider with the development of an assistive interface is perceived ease of use~\cite{marangunic2015technology, venkatesh2000determinants}. Participants noted the continuous control of the robot using the head-worn interface which allowed for intuitively varying the velocity of the robot's actuators in proportion to the magnitude of head tilt. When participants were presented with the statement ``The control interface was easy to use'' (see Fig.~\ref{finlik}), healthy participants agreed with a median ranking of 6 while participants with impairments agreed with a median ranking of 6.5 out of 7. Additionally, in the qualitative interviews, 9$/$16 healthy participants and both participants with impairments mentioned, without solicitation, that control using the head-worn was intuitive or more intuitive than the web interface. I2 likened the interface to a Segway, ``If you wanted to go forward, you would pitch forward... so I would say strangely the hat was more intuitive.''

\subsection{Learning Curve}
\begin{figure}[t!]
  \centering
  \includegraphics[width = \columnwidth]{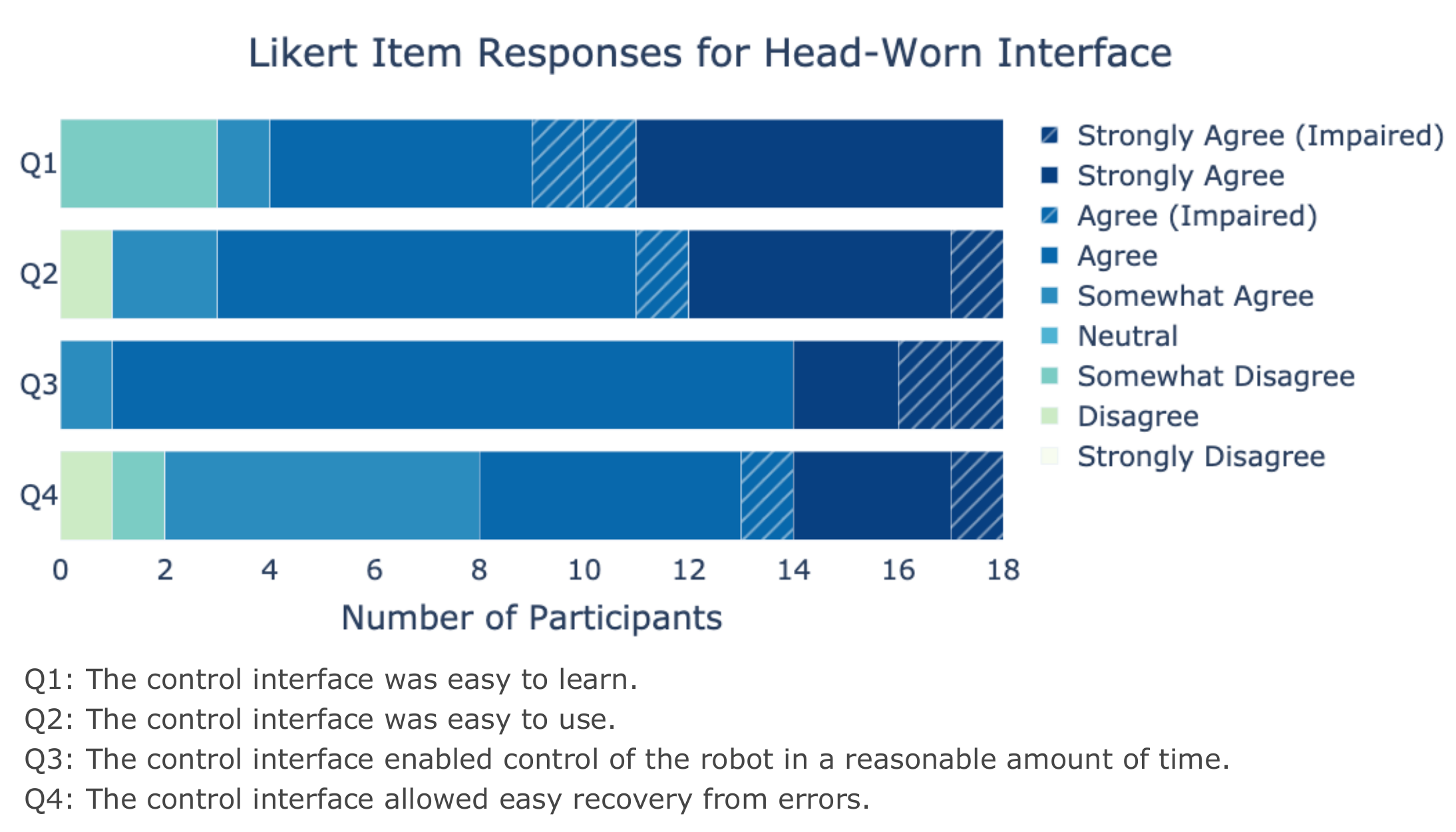}
  \vspace{-0.5cm}
  \caption{Results from the 7-point Likert items presented to participants after completion of all tasks using the inertial head-worn interface. The majority of participants agree with all four statements about the proposed interface.}
  \vspace{-0.3cm}
  \label{finlik}
\end{figure}

For the head-worn interface, both healthy participants and participants with impairments agreed that ``The control interface was easy to learn'', as seen in Fig.~\ref{finlik}, with a median reported score of 6 (Agree) for both groups. 4 out of the 16 healthy participants felt that the web interface was faster to learn, but mentioned that use of the head-worn interface improved with more practice. One participant (P8) noted at the end of the study: ``If I had never seen any of this before and you wanted me to do one thing and that was it, I would take the web interface but with time to practice, the hat was definitely way better''. 

\subsection{Workload}

\begin{figure}[t!]
  \centering
  \includegraphics[width = \columnwidth]{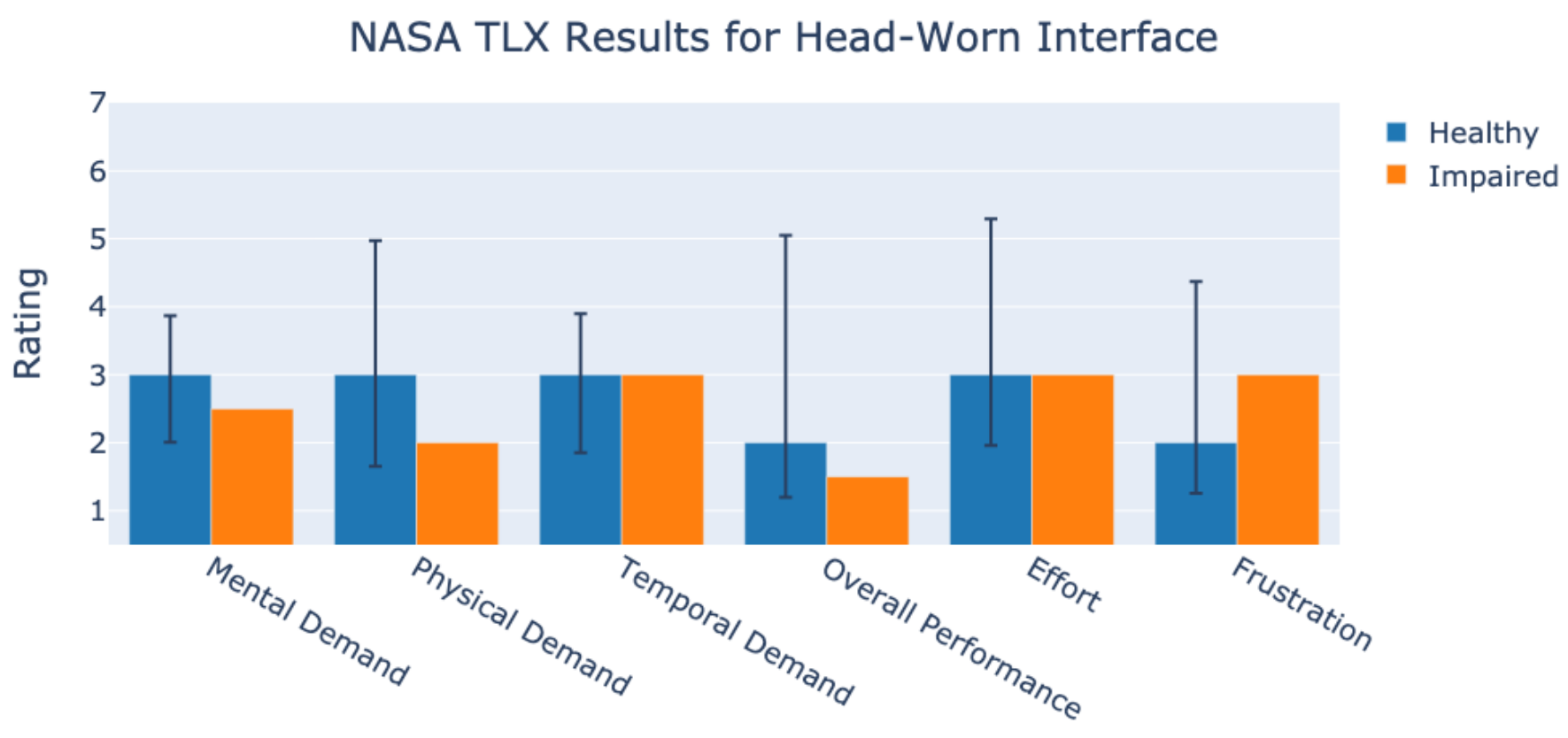}
  \vspace{-0.5cm}
  \caption{Median results from the NASA TLX survey asked to participants after completion of all tasks using the head-worn interface. A rating of 1 is best for all categories.}
  \vspace{-0.5cm}
  \label{nasatlx}
\end{figure}
Workload and perceived effort are also important metrics to evaluate for new interfaces~\cite{gopher1986workload, meijman2013psychological}. The NASA TLX~\cite{hart2006nasa} is a 7-point scale that measures mental, physical, and temporal demand, overall performance, effort, and frustration. A rating of 1 is best for all categories. The median results are located in Fig.~\ref{nasatlx}. 
The overall performance of the interface was strong for both healthy participants and participants with impairments with a median ranking of 2 and 1.5 respectively. 

For mental demand, healthy participants and participants with impairments rated the head-worn interface as 3 and 2.5 respectively. 
However, 5 healthy participants mentioned facing mental demand as the head-worn interface required them to recall more commands in contrast to having them displayed on the computer screen in the web interface. Additionally, both participants with impairments mentioned demand associated with remembering to keep their head still while not trying to command robot motions. For physical demand, healthy participants and participants with impairments rated the head-worn interface as low with 3 and 2 respectively. 

\subsection{Challenges and Future Directions}
Perception was a challenge for participants while using the head-worn interface. In Task 1, the lack of depth perception was an issue due to the positioning of the cup far away on a kitchen counter. In qualitative interviews, 11$/$16 healthy participants highlighted that the camera feed of the web interface helped with this problem and a few vocalized that a combination of the head-worn interface and the visual cues from the web interface would work the best. In the qualitative interview, 6 healthy participants and 1 participant with impairments mentioned preferring interaction directly with the robot through line of sight. One participant (P14) said, ``When I'm trying to operate the robot with [the head-worn interface], I'm still a participant in the scene, experiencing reality, but when I'm looking at the screen that's taking all my attention.'' In addition, having some visual interface with a camera feed is necessary in situations where the robot is used out of line of sight, such as in another room. Participants also mentioned issues with maintaining vision of objects while tilting their head. This was specifically a problem while closing the gripper which required participants to tilt their head up, restricting line of sight to objects like the blanket or towel that the robot was manipulating near the person. 

A few participants also expressed a desire for more visual feedback with the head-worn interface, specifically a diagram to remind them of the direction to tilt their head and to show them where the thresholds were. Future work could explore providing participants with these visual cues. 
Several participants faced occasional difficulty using speech recognition, which presents a future opportunity to provide users with other methods for mode switching.
Both participants with motor impairments also mentioned issues with depth perception and mode switching. I1 additionally mentioned issues with his impairment limiting head movement in the roll (Y) axis, preventing him from going as fast as he would have liked. He elaborated that a calibration procedure could help make the tilting range for the roll axis smaller, allowing him to reach faster speeds.

\section{CONCLUSIONS}
This work presents a unique inertial-based wearable assistive interface, embedded in a familiar head-worn garment, for individuals with motor impairments to teleoperate and perform physical tasks with a mobile manipulator. The interface presents individuals with impairments an alternative to conventional computer based teleoperation platforms. Our results from a study with both able-bodied and participants with impairments show that the head-worn interface is both intuitive, and efficient, allowing users to perform a variety of physical self-care and household tasks with few errors. Our study results contrast the head-worn interface with a conventional computer-based assistive web interface and we observed benefits in terms of efficiency and ease of use.






\bibliographystyle{IEEEtran}
\bibliography{bibliography}

\end{document}